\documentclass{article}
\usepackage{ijcai24}

\usepackage{times}
\usepackage{soul}
\usepackage{url}
\usepackage[hidelinks]{hyperref}
\usepackage[utf8]{inputenc}
\usepackage[small]{caption}
\usepackage{graphicx}
\usepackage{amsmath}
\usepackage{amsthm}
\usepackage{booktabs}
\usepackage{algorithm}
\usepackage{algorithmic}
\usepackage[switch]{lineno}

\usepackage{booktabs}
\usepackage{multirow}
\usepackage{amssymb}
\usepackage{amsmath}
\usepackage{color}

\usepackage[marginal]{footmisc}

\title{ZeroDDI: A Zero-Shot Drug-Drug Interaction Event Prediction Method with Semantic Enhanced Learning and Dual-Modal Uniform Alignment}

\author{
Ziyan Wang $^{\dag}$
 \and
Zhankun Xiong $^{\dag}$ \and
Feng Huang \and
Xuan Liu\And
Wen Zhang $^{*}$  \\
\affiliations
 College of Informatics, Huazhong Agricultural University, Wuhan 430070, China\\
\emails
\{wangziyan, xiongzk, fhuang233, lx666\}@webmail.hzau.edu.cn,
zhangwen@mail.hzau.edu.cn
}

\begin{document}

\maketitle

\begin{abstract}
Drug-drug interactions (DDIs) can result in various pharmacological changes, which can be categorized into different classes known as DDI events (DDIEs). In recent years, previously unobserved/unseen DDIEs have been emerging, posing a new classification task when unseen classes have no labelled instances in the training stage, which is formulated as a zero-shot DDIE prediction (ZS-DDIE) task. However, existing computational methods are not directly applicable to ZS-DDIE, which has two primary challenges: obtaining suitable DDIE representations and handling the class imbalance issue. To overcome these challenges, we propose a novel method named ZeroDDI for the ZS-DDIE task. Specifically, we design a biological semantic enhanced DDIE representation learning module, which emphasizes the key biological semantics and distills discriminative molecular substructure-related semantics for DDIE representation learning. Furthermore, we propose a dual-modal uniform alignment strategy to distribute drug pair representations and DDIE semantic representations uniformly in a unit sphere and align the matched ones, which can mitigate the issue of class imbalance. Extensive experiments showed that ZeroDDI surpasses the baselines and indicate that it is a promising tool for detecting unseen DDIEs. Our code has been released in https://github.com/wzy-Sarah/ZeroDDI.
\end{abstract}

\section{Introduction}

\begin{figure}[!t]\centering
	\includegraphics[width=8.5cm]{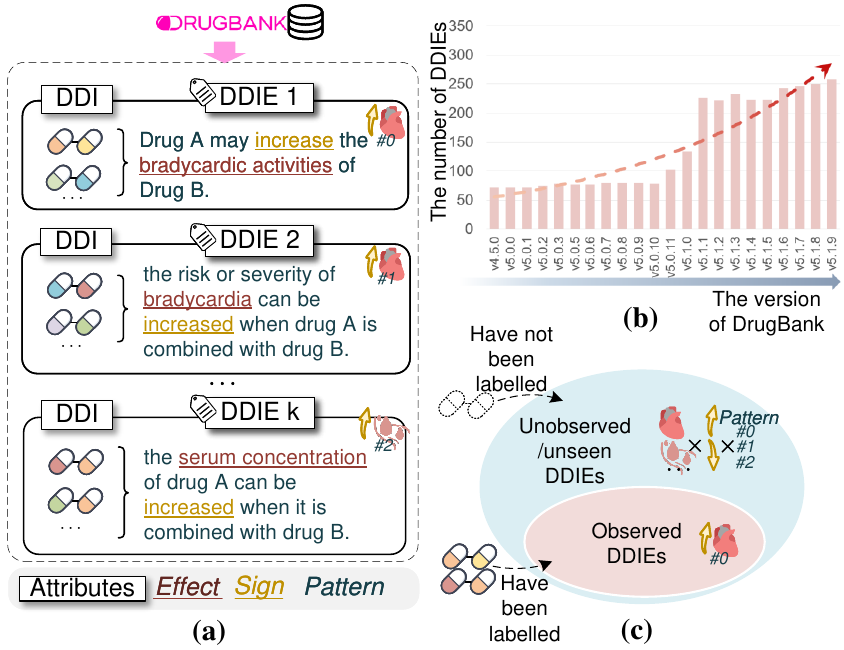}
	\caption{ (a) DrugBank is a standard database containing the textual descriptions of DDIEs. Almost every description can be split into \emph{Effect}, \emph{Sign}, and \emph{Pattern} attributes. (b) With the updates of the Drugbank version, the number of DDIEs is also increasing. (c) Unseen DDIEs may come from the composition of existing attributes.}
    \label{FIG_1}
\end{figure}

Taking multiple drugs concurrently can cause drug-drug interactions (DDIs), which may lead to various pharmacological changes  \cite{vilar_similarity-based_2014}. According to the different specific consequences, the huge number of DDIs can be categorized into hundreds of classes \cite{ryu_deep_2018}, named DDI events (DDIEs), as shown in Figure \ref{FIG_1}(a). In recent years, researchers have paid much attention to DDIE prediction, i.e., classifying a DDI into a specific DDIE, which is helpful for researchers and clinicians to investigate the mechanism behind the consequences of polypharmacy. Note that to avoid confusion, we will use the term "drug pairs" instead of DDIs in this paper.

With further clinical observation and the development of therapeutic drugs, the number of DDIEs is increasing, as shown in Figure \ref{FIG_1} (b). Moreover, inspired by \cite{deng2020,feng_social_2023}, we notice that the DDIE contains three key concepts: the effect (such as “bradycardia”), the “increased” or "decreased" pharmacological changes, and the directional interaction between two drugs (which is reflected in the sentence pattern). To simplify, we term them as three kinds of attributes, i.e., \emph{Effect}, \emph{Sign}, and \emph{Pattern}, respectively. We observed that annotated DDIEs (seen DDIEs) are less than all compositions of three kinds of attributes, as shown in Figure \ref{FIG_1} (c), thus we can infer the existence of unseen DDIEs without labelled drug pairs. Predicting unseen DDIEs has a high practical value because novel drug pairs may come from unseen DDIEs. We formulate classifying drug pairs of unseen DDIEs as a zero-shot DDIE prediction (ZS-DDIE) task.

Although many DDIE prediction methods have been proposed, they all neglect the ZS-DDIE task. When no trainable instances are used to transfer knowledge from seen to unseen classes, the semantic information (or auxiliary information) is needed to be attached to every class for constructing semantic relatedness between seen and unseen classes. However, most DDIE prediction methods view DDIEs as either labels \cite{zhu_learning_2023}, random initialized vectors \cite{lin_r2-ddi_2022} or one-hot vectors \cite{zhu_molecular_2022}, which do not have realistic semantic significance. Therefore, they cannot be applied to the ZS-DDIE task. An intuitive solution for ZS-DDIE is to learn the experience from zero-shot learning (ZSL) classification methods, which are developed by computer vision. There is a commonly used compatibility framework in ZSL \cite{xian_zero-shot_2018}, which transfers knowledge from seen to unseen classes by projecting all classes into a common semantic space, and realizes the unseen class prediction by aligning the representations of instances and matched classes. Nevertheless, applying this framework to the ZS-DDIE task needs to resolve two major challenges. \textbf{The first challenge} is to obtain suitable DDIE representations, which are required to construct reasonable relatedness between different classes. For example, based on DDIE textual descriptions, the DDIE representations with the same \emph{Pattern} may be closer than that with the same \emph{Effect} in semantic space (as shown in Appendix A), which is inconsistent with the biological significance and will hinder the prediction performance. \textbf{The second challenge} is the class imbalance. Researchers have revealed that the top three most frequent DDIEs contain more than half of all instances \cite{deng_meta-ddie_2021}, while a great number of DDIEs have less than 10 instances. The class imbalance problem may lead to unclear decision boundaries of classification, causing a decline of discriminative ability for the prediction of classes with few or even no instances \cite{imbalance}.

In this work, we propose a novel method, called ZeroDDI, for zero-shot DDIE prediction and resolve the challenges above. \textbf{For the first challenge}, we design a biological semantic enhanced DDIE representation learning module (BRL) to obtain suitable DDIE representation. Since \emph{Effect} attributes are the key biological semantics, BRL first extracts attribute-level semantics of \emph{Effect} and together with class-level semantics of DDIE textual descriptions. Then, considering that molecular substructure plays an important role in drug properties \cite{jin_general_2023}, BRL establishes the fine-gained interaction between substructures and semantic tokens of text to guide the fusion of bi-level semantics, which can distill discriminative semantics for learning DDIE representations. \textbf{For the second challenge}, we design a dual-modal uniform alignment strategy (DUA) to enable the representations of both drug pairs (structure modal) and DDIEs (text modal) uniformly distributed on the unit sphere and then align the matched ones, which can mitigate the problem of unclear decision boundaries of classification that caused by class imbalance \cite{li_targeted_2022} and further improve the ZS-DDIE performance.

In summary, the main contributions of this paper are described as follows:
\begin{itemize}

\item  We investigate a novel problem for DDIE prediction: zero-shot DDIE prediction (ZS-DDIE), and propose a novel method ZeroDDI for the ZS-DDIE task, along with a ZS-DDIE dataset.

\item  We design a novel biological semantic enhanced DDIE representation learning module (BRL) with the extraction of class-level and attribute-level semantics, and substructure-guided bi-level semantic fusion to learn suitable DDIE representations with reasonable relatedness between different classes.

\item We introduce a dual-modal uniform alignment strategy (DUA) to enable structure and text modal representations (i.e., drug pair and DDIE representations) to be uniformly distributed in unit sphere, aiming to handle the class imbalance issue and thus improve the discrimination of the model.

\item The extensive experiments show that ZeroDDI can achieve superior performances in the ZS-DDIE task than baselines and reveal that ZeroDDI is a promising tool for detecting unseen DDIEs.

\end{itemize}

\section{Related Work}

\subsection{Drug-Drug Interaction Event Prediction}

Existing DDIE prediction methods can be classified into three categories: DNN-based methods \cite{deng_meta-ddie_2021,lin2021mdf}, Tensor factorization-based methods \cite{jin_multitask_2017,yu_stnn-ddi_2022}, and GNN-based methods \cite{zhu_molecular_2022,lv_3d_2023,xiong2023multi}. However, none of the DDIE prediction methods are specifically tailored to tackle the ZS-DDIE task. Most of them cannot be applied to the ZS-DDIE task in that they neglect to encode DDIE or lack semantic significance in their DDIE embeddings \cite{zhu_learning_2023,zhu_molecular_2022,lin_r2-ddi_2022}. We find that 3DGT-DDI \cite{lv_3d_2023}, which utilized a pre-trained language model (SCIBERT) to learn DDIE representations from textual descriptions, shows potential adaptability for handling the ZS-DDIE task. Despite this, zero-shot DDIE prediction remains uncharted territory and needs to be explored. More discussions of related work are in Appendix B.

\begin{figure*}[!t]\centering
	\includegraphics[width=18.2cm]{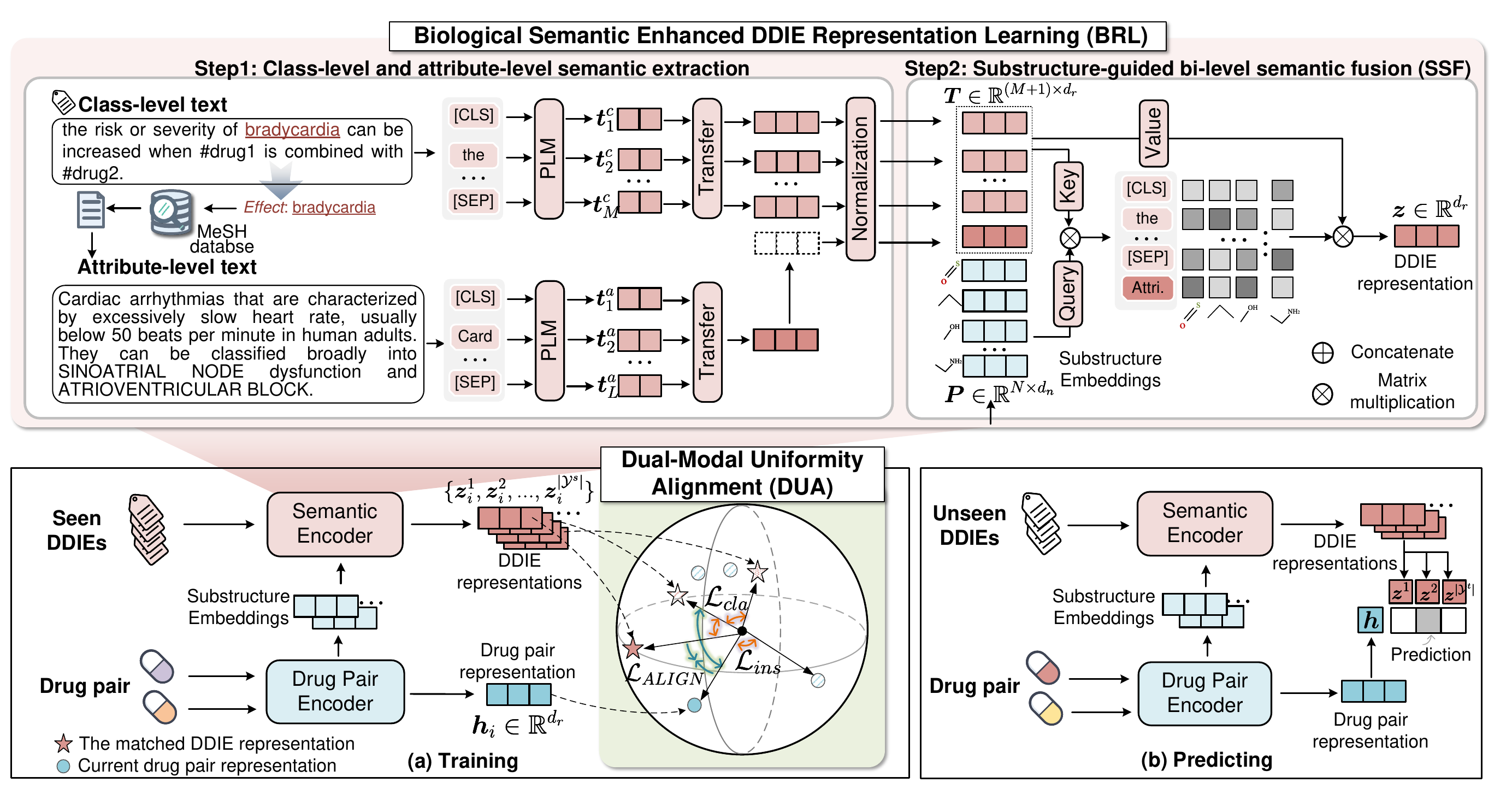}
	\caption{The overall framework of ZeroDDI.}
    \label{overview}
\end{figure*}

\subsection{Zero-Shot Learning}

Zero-Shot learning (ZSL) methods, which were introduced in computer vision, achieve transferability by projecting all classes into a common semantic space, and facilitate the classification of instances from unseen classes through a compatibility function \cite{nawaz_semantically_2022}. For learning semantic representation, current ZSL methods rely on three primary information sources: class names, class attributes, and class textual descriptions. Word-based methods \cite{frome_devise_2013,li_zero-shot_2015} represent the classes by simple words of the class name. Attribute-based methods represent the classes by more accurate and specific shared characteristics among classes \cite{lampert_attribute-based_2013,al-halah_automatic_2017}. Textual description-based methods represent the classes by textual descriptions of classes \cite{reed_learning_2016,zhu_generative_2018}, which can provide more context information about classes. For constructing compatibility functions, ZSL methods utilized/developed different metric functions to pull an instance close to its matched class and push it away from other classes. For example,  DUET \cite{chenduet2023}, Sumbul et al. \cite{sumbul_fine-grained_2017} use cross-entropy loss to enforce the instance to have the highest compatibility score with its matched class; DeViSE \cite{frome_devise_2013}, ALE \cite{ALE} and Cacheux et al. \cite{tripletloss} use rank loss to push the matched pairs and the mismatched pairs away by a margin.

In the ZS-DDIE task, although textual descriptions contain more context information, the key semantics are not emphasized, thus we add \emph{Effect} attribute semantics to enhance the biological semantic. Moreover, inspired by the success of CLIP \cite{CLIP}, we utilize a contrastive loss to maximize the similarity of a drug pair and its matched DDIE for classification.

\section{Methodology}

\subsection{Preliminaries}
Given the training data $\mathcal{S}=\left \{ ((g_{i},g^{\prime}_{i}), y_{i})\right\}_{i=1}^{I}$, where each element includes a pair of drugs $(g_{i},g^{\prime}_{i})$ as an instance and a matched DDIE $y_i \in \mathcal{Y}^{s}$ as the label, and $\mathcal{Y}^{s}$ denotes the set of seen DDIE labels. Let $\mathcal{Y}^{u}$ denote the set of unseen DDIE labels, which is disjoint from $\mathcal{Y}^{s}$, i.e. $\mathcal{Y}^{s} \cap \mathcal{Y}^{u}=\emptyset$.  We denote $\mathcal{T}=\left \{ ((g_{j}, g^{\prime}_{j}),y_{j})\right\}_{j=1}^{J}$ as the testing data and $\mathcal{Y}^{t}$ as the set of test DDIE labels, $y_j \in \mathcal{Y}^{t}$. Herein we test the model in two scenarios: one is that $\mathcal{Y}^{t}=\mathcal{Y}^{u}$, i.e., all test classes come from the unseen class set, named conventional ZSL (CZSL); the other is that $ \mathcal{Y}^{t}=\mathcal{\Tilde{Y}}^{s}\cup\mathcal{Y}^{u}, \mathcal{\Tilde{Y}}^{s} \subseteq \mathcal{Y}^{s}$, i.e., the test classes come from both seen class set and unseen class set, named generalized ZSL (GZSL). GZSL is a more realistic scenario where the model lacks information about whether a novel instance belongs to the seen or unseen class. The goal of the ZS-DDIE task is to leverage the training data $\mathcal{S}$ from seen classes to learn a model that predicts the label of each instance in $\mathcal{T}$, which is a multi-class classification task.

\subsection{Overview}
The architecture of ZeroDDI is shown in Figure \ref{overview}. Based on the compatibility framework, we first obtain the drug pair representations and molecular substructure embeddings of drug pairs from a Drug Pair Encoder (based on \cite{MSAN} and details can be seen in Appendix C). Then we get the DDIE representations by our designed BRL. Thereafter, DUA is designed to align the drug pair representations and their matched DDIE representations in unit sphere space and train the model. Finally, in the prediction phase, the unseen class whose DDIE representation has the maximum dot-product similarity score with the drug pair is the prediction. 


\subsection{Biological Semantic Enhanced DDIE Representation Learning (BRL)}
To obtain DDIE representations, we design a BRL module, which is composed of two steps: class-level and attribute-level semantic extraction, and substructure-guided bi-level semantic fusion (SSF). The former extracts the bi-level semantics of DDIEs from the attribute-level and class-level text. The latter adaptive fuses the bi-level semantic to obtain DDIE representations, preserving discriminative information as much as possible.

\paragraph{Class-level and attribute-level semantic extraction.} For the class-level texts, we extract the DDIE textual descriptions from the DrugBank database \cite{wishart2018drugbank}. For the attribute-level texts, we utilize a StanfordNLP tool \cite{qi2019universal} to locate the \emph{Effect} words in every DDIE textual description, and we extract the textual description of the \emph{Effect} from the MeSH database\footnote{MeSH (Medical Subject Headings) is the NLM controlled vocabulary thesaurus used for indexing articles for PubMed}. After that, we employ a language model BioBERT \cite{lee_biobert_2020} pre-trained on large-scale biomedical corpora, to get the token features $\left \{\boldsymbol{t}^{c}_{m} \in \mathbb{R}^{d_{t}} \right \}_{m=1}^M$ and $\left \{\boldsymbol{t}_{l}^{a} \in \mathbb{R}^{d_{t}} \right \}_{l=1}^L$ from class-level text and attribute-level text, respectively, where $M$ and $L$ are the numbers of the tokens and $d_{t}$ denote the token feature dimension. Since the attribute is part of a DDIE, we take the average of attribute-level token features as extra token information attached to class-level token features. Formally, we can obtain the bi-level token-wise feature $\boldsymbol{T} \in \mathbb{R}^{(M+1)\times d_{r}}$ by:
\begin{equation}
    \boldsymbol{T}=LN\left [\phi_{1}\left( \left [\boldsymbol{t}^{c}_{1};\boldsymbol{t}^{c}_{2};...;\boldsymbol{t}^{c}_{M}  \right ]\right); \frac{1}{L} \sum_{l}^{L} \phi_{2}(\boldsymbol{t}_{l}^{a}) \right ]
\end{equation}
\noindent where $\phi_{1}$ and $\phi_{2}$ are MLPs transferring $\mathbb{R}^{d_{t}} \rightarrow \mathbb{R}^{d_{r}}$ to reduce the token feature dimension to $d_{r}$. The symbol $;$ denotes concatenate. $LN$ is Layer Normalization \cite{ba2016layer}, which is used to normalize the features that come from two-level semantic sources. 

\paragraph{Substructure-guided bi-level semantic fusion (SSF).} Different from averaging the tokens $\boldsymbol{T}$ directly, we utilize a cross-modality attention mechanism to fuse the tokens discriminatively, i.e., $\mathbb{R}^{(M+1) \times d_{r}} \rightarrow \mathbb{R}^{1 \times d_{r}}$. In particular, we take the molecular substructure embeddings of drug pair $\boldsymbol{P} \in \mathbb{R}^{N \times d_{n}}$ (learned from Drug Pair Encoder) to establish the fine-grained interaction with $\boldsymbol{T}$ and distill the substructure-related token semantics, where $N$ is the number of substructures. More specifically, $\boldsymbol{P}$ and $\boldsymbol{T}$ are transferred to: 
\begin{equation}
\boldsymbol{Q}=\boldsymbol{P}\boldsymbol{W}_{Q}, \boldsymbol{K}=\boldsymbol{T}\boldsymbol{W}_{K},  \boldsymbol{V}=\boldsymbol{T} \boldsymbol{W}_{V}
\end{equation}
\noindent where $\boldsymbol{Q} 
\in \mathbb{R}^{N \times d_{r}}$ denotes the queries, $\boldsymbol{K} \in \mathbb{R}^{(M+1) \times d_{r}}$ and $\boldsymbol{V} \in \mathbb{R}^{(M+1) \times d_{r}}$ denote keys and values. $\boldsymbol{W}_{Q}$, $\boldsymbol{W}_{K}$ and $\boldsymbol{W}_{V}$ are learnable linear transformations. Then the cross-modality attention can be calculated by:
\begin{equation}
\boldsymbol{A}=\operatorname{softmax}\left(\frac{\boldsymbol{Q} \boldsymbol{K}^T}{\sqrt{r}}\right), \boldsymbol{O}=\boldsymbol{A} \boldsymbol{V},
\end{equation}
\noindent where $\boldsymbol{A} \in \mathbb{R}^{N \times (M+1)}$, $\boldsymbol{O} \in \mathbb{R}^{N \times d_{r}}$. Finally, averaging over all the rows of $\boldsymbol{O}$, we can obtain the DDIE representation $\boldsymbol{z} \in \mathbb{R}^{d_{r}}$.

\subsection{Dual-Modal Uniform Alignment (DUA)} 
We introduce a DUA as the compatibility function, which includes an alignment function to align the drug pair representations and its matched DDIE representations for classification, and a dual-modal uniformity loss to constraint both modal representations uniformly distributed in unit sphere.

To realize the classification of drug pairs, we use the dot-product to measure the similarity of the drug pair representation with all DDIE semantic representations and then employ contrastive loss \cite{han_contrastive_2021} to enforce the drug pair to have the highest similarity score with its matched DDIE: 

\begin{equation}
    \mathcal{L}_{ALIGN}=-\sum_{i=1}^I\log \frac{\exp \left(\boldsymbol{h}_i \cdot \boldsymbol{z}^{y_{i}}_i / \tau\right)}{ \sum_{j \in\mathcal{Y}^{s}} \exp (\boldsymbol{h}_i \cdot \boldsymbol{z}^j_i / \tau )}
\end{equation}

\noindent where $\boldsymbol{h}_i$ denotes the current drug pair representation, $\boldsymbol{z}^{y_i}_i$ denotes the matched DDIE representation of $\boldsymbol{h}_i$, $\boldsymbol{z}^j_i \in \{\boldsymbol{z}^1_i,\boldsymbol{z}^2_i,...,\boldsymbol{z}^{|\mathcal{Y}^{s}|}_i\}$  denotes the $j$-th DDIE representations of all DDIE representations corresponding to $\boldsymbol{h}_i$. $\tau$ is a temperature coefficient.

To realize the uniform distribution, i.e., maximize the representation distance between classes in unit sphere space, inspired by \cite{lu_geometer_2022}, we take the center of all the DDIE representations as the center of the sphere. Then we utilize the class uniformity loss $\mathcal{L}_{cla}$ to pull every DDIE representation to unit sphere:

\begin{equation}
    \resizebox{1.0\linewidth}{!}{$
            \displaystyle
                \mathcal{L}_{cla}=\frac{1}{I |\mathcal{Y}^{s}|} \sum_{i=1}^{I}  \sum_{j=1}^{|\mathcal{Y}^{s}|}\left\{1 +
   \max _{k \in \mathcal{Y}^{s} \backslash \{j\}}\left[\frac{(\boldsymbol{z}_i^j-\boldsymbol{c}_{i})}{\|\boldsymbol{z}_i^j-\boldsymbol{c}_{i}\|} \cdot  \frac{(\boldsymbol{z}_i^k-\boldsymbol{c}_{i})}{\|\boldsymbol{z}_i^k-\boldsymbol{c}_{i}\|}\right]\right\}
        $}
\end{equation}

\noindent where $\boldsymbol{c}_{i}=1/| \mathcal{Y}^{s}| \cdot \sum_{j=1}^{|\mathcal{Y}^{s}|} \boldsymbol{z}^j_{i}$ and denotes the center of all DDIE representations of drug pair $i$. $k \in\mathcal{Y}^{s} \backslash \left\{j\right\}$ means sampling all DDIEs except the $j$-th DDIE. This loss function uses the cosine similarity to pull every distance between the DDIE representations and the center representation approaching to same. Thus all DDIE representations are unified and distributed on a unit sphere. 

To make the drug pair representation get consistency constraints with DDIE representations. We introduce an instance uniformity loss $\mathcal{L}_{ins}$ to unify every drug pair $i$ to the sphere with the center of $\boldsymbol{c}_{i}$:

\begin{equation}
\mathcal{L}_{ins}=\frac{1}{I} \sum_{i=1}^{I}\left\{1+\max _{k \in \mathcal{S} \backslash \{i\}}\left[\frac{\left(\boldsymbol{h}_{i}-\boldsymbol{c}_{i}\right)}{\left\|\boldsymbol{h}_{i}-\boldsymbol{c}_{i}\right\|} \cdot \frac{\left(\boldsymbol{h}_{k}-\boldsymbol{c}_{i}\right)}{\left\|\boldsymbol{h}_{k}-\boldsymbol{c}_{i}\right\|}\right]\right\}
\end{equation}
where $ k \in \mathcal{S} \backslash \left\{i\right\}$ is the sample randomly drawn from a batch set of training data excluding $i$. According to Eq. (5) and Eq. (6), the dual-modal uniformity loss $\mathcal{L}_{UNI}$ can be expressed as:

\begin{equation}
\mathcal{L}_{UNI}=\mathcal{L}_{cla}+\mathcal{L}_{ins}
\end{equation}

\subsection{Training and Prediction}
For training our model ZeroDDI, we optimize the total loss that combines Eq. (4) and Eq. (7):

\begin{equation}
\mathcal{L}=\mathcal{L}_{ALIGN}+\lambda \mathcal{L}_{UNI}
\end{equation}

\noindent where $\lambda$ is hyper-parameters.

For the zero-shot DDIE prediction, we use the trained Drug Pair Encoder to obtain drug pair representation $\boldsymbol{h}$. Its DDIE $\hat{y}$ can be predicted by researching the highest dot-product similarity between $\boldsymbol{h}$ and all DDIE representations in test set:
\begin{equation}
\hat{y}=\underset{j \in \mathcal{Y}^{t}}{\operatorname{argmax}} \,\boldsymbol{h} \cdot \boldsymbol{z}^j,
\end{equation}
where $\mathcal{Y}^{t}$ is the set of test DDIEs, $\boldsymbol{z}^j$ denotes the $j$-th DDIE representation learned by the trained BRL.

\section{Experiments}

\subsection{Experimental Setup}
\subsubsection{Datasets} DrugBank \cite{wishart2018drugbank} is a widely used data source for DDIE prediction, which contains drugs, DDIs, textual descriptions of DDIEs, and the molecular structures of drugs. Based on raw data from DrugBank v5.1.9, we construct a ZS-DDIE dataset by annotating attributes for every DDI manually, which contains a total of 2,004 approved drugs, 394,118 DDIs, 175 DDIEs with unique textual descriptions, 2 \emph{Signs}, 3 \emph{Patterns} and 114 \emph{Effects} with corresponding attribute-level text. We rank the classes of DDIEs in the order of the number of instances from more to less and split DDIEs into seen classes and unseen classes, in which we take the less number of DDIEs as unseen classes to simulate the real situation of DDIE that are newly observed. 107 DDIEs are viewed as seen classes $\mathcal{Y}^{s}$, which contains all attributes, and the rest 68 DDIEs are viewed as unseen classes $\mathcal{Y}^{u}$. Additionally, every DDI is associated with a DDIE in this dataset. Drug pairs in the dataset are asymmetric, in that the roles of drugs are different in one DDIE \cite{feng_social_2023}. More details of the dataset can be found in Appendix D.1.

\subsubsection{Baselines} As mentioned in the related work, only one DDIE method (3DGT-DDI) has the potential to be applied to the ZS-DDIE task. In addition, under the compatibility framework, we also compare several popular compatibility functions in computer vision and equip them with two kinds of DDIE semantic representations. One learned from a binary vector of attributes (namely attribute-based representation) and the other learned from a PLM embedding of class textual description (namely class-based representation), respectively. These baselines have the same Drug Pair Encoder as our method.

\begin{itemize}

\item \textit{3DGT-DDI} \cite{lv_3d_2023}. The DDIE semantic encoder of 3DGT-DDI is composed of a PLM and a CNN. To enable this model to deal with the ZS-DDIE task, we replace its binary classifier with a compatibility function incorporating cross-entropy loss.

\item \textit{ZSLHinge}. ZSLHinge utilizes a hinge rank loss from the classic ZSL method DeViSE \cite{frome_devise_2013}, which can produce a higher dot-product similarity between an instance and the matched class than between the instance and other randomly chosen classes by a margin. 

\item \textit{ZSLCE}. ZSLCE utilizes the cross-entropy loss from \cite{chenduet2023} and \cite{sumbul_fine-grained_2017}, which maximizes the similarity score of the instance and its matching class.

\item \textit{ZSLTriplet}. ZSLTriplet utilizes a ZSL triplet loss from \cite{tripletloss}, which improves the standard triplet loss by adding a flexible semantic margin, partial normalization, and relevance weighting.

\end{itemize}

More details of baselines can be found in Appendix D.2. 

\begin{table*}[htbp]
\centering
\setlength{\tabcolsep}{2.1mm}
{\begin{tabular}{ccccccccccccc}
\toprule
\multirow{2}{*}{Model}& \multicolumn{4}{c}{CZSL} & \multicolumn{8}{c}{GZSL} \\ \cmidrule(l){2-5} \cmidrule(l){6-13} 
 & 
$\textbf{acc}_{ave}^{u}$ &
$\textbf{acc}_{@1}^{u}$  & $\textbf{acc}_{@3}^{u}$ & $\textbf{acc}_{@5}^{u}$ & 
$\textbf{acc}^{u}_{ave}$ & 
$\textbf{acc}_{@1}^{u}$ &
$\textbf{acc}^{s}_{ave}$ &
$\textbf{acc}_{@1}^{s}$ &
$\textbf{acc}^{H}_{ave}$ &
$\textbf{acc}_{@1}^{H}$  &
 $\textbf{P}^{u}$     &
 $\textbf{P}^{s}$

   \\ \midrule

3DGT-DDI&8.25 & 13.50 &27.93 & 39.27 & 8.21&12.55 &48.30&45.29&14.03&19.65& 14.89&90.15\\

ZSLHinge$^{1}$& 5.47 &5.75& 21.17 &30.45 &5.39 &5.69 &23.14 & 22.65&8.06 & 9.09& 5.87&\textbf{98.05}\\

ZSLHinge$^{2}$&12.80 & 10.69& 28.74&42.46 & 10.01&10.1 &54.93&53.02 &16.87   &15.86& 11.08&95.48 \\

ZSLCE$^{1}$  &11.35 &11.32 & 30.21& 44.72&7.55 &8.90 & 54.57&61.75 &12.97 & 15.14 & 13.65&75.19 \\

ZSLCE$^{2}$  &15.08 & 16.86&39.04 & 53.34& \underline{12.73}& 14.84&76.20 & 71.03&\underline{21.78} & 24.27 & 17.76&  97.01\\


ZSLTriplet$^{1}$  & \underline{15.09}& 14.77&40.51& \underline{59.84}& 10.23& 10.92&\textbf{78.71} &\textbf{74.20 }&18.05 &18.61 &15.09& 94.15\\

ZSLTriplet$^{2}$  &13.51 &18.80 & 28.36& 51.27& 11.75&\underline{16.22} &66.48 & 62.37&19.89 & 20.19&19.13&  96.25 \\

ZeroDDI$^{1}$& 13.16  &  15.79& 40.04 & 52.94&10.65& 13.31 & 72.70 & 69.35 & 18.19 & 20.95&15.48 &\underline{97.87}\\
 
 ZeroDDI$^{2}$&  14.66&\underline{19.22}  &\underline{42.65} &56.28&11.72&15.10 & 74.09 & \underline{73.66} &  20.21& \underline{24.83}&\underline{19.42}&96.93 \\
 
ZeroDDI & \textbf{17.57} & \textbf{21.35} & \textbf{47.81} & \textbf{66.16}&\textbf{16.13}& \textbf{19.12} & \underline{77.25} & 70.67 & \textbf{26.48}  & \textbf{29.14}&\textbf{22.09}&97.02 \\
    \bottomrule
\end{tabular}}%

\caption{Performance (in \%) comparisons of ZeroDDI with baselines in the CZSL and GZSL scenario. The best and suboptimal results are highlighted in \textbf{bold} and \underline{underline}, respectively. Note that Model$^{1}$ (or Model$^{2}$) denotes using attribute-based (or class-based) representations learned from a binary vector of attributes (or the PLM of class textual description where the semantic encoder is the same as that in ZeroDDI.) }
\label{baselines}
\end{table*}

\subsubsection{Evaluation Protocols} 
The experiments are performed on two common zero-shot learning scenarios: CZSL and GZSL, which are described in Preliminaries. We train the model on data from seen classes.
$\textbf{In the CZSL scenario}$, we validate and test the model in 3-fold of the unseen classes, i.e., in each fold, one-third of unseen classes are used for the validation, and the rest are used for the testing. $\textbf{In the GZSL scenario}$, to evaluate the performance of seen and unseen prediction simultaneously, a part of the seen classes and instances are added into the GZSL validation and testing set. Note that all instances from seen classes in the validation and testing set are not included in the training set.  

\subsubsection{Metrics}
In the CZSL scenario, the performance is measured with top-k accuracy (k=1, 3, 5 and denoted as $\textbf{acc}_{@1}^{u}$, $\textbf{acc}_{@3}^{u}$ and $\textbf{acc}_{@5}^{u}$). In addition, we calculate the average of top-1 accuracy across all unseen classes, denoted as $\textbf{acc}_{ave}^{u}$. In the GZSL scenario, we calculate the $\textbf{acc}_{@1}$ or $\textbf{acc}_{ave}$  on seen classes (denoted as $\textbf{acc}^{s}_{@1}$ or $\textbf{acc}^{s}_{ave}$) and unseen classes, and calculate the harmonic mean of them: $\textbf{acc}^{H}=2*(\textbf{acc}^{s} * \textbf{acc}^{u})/(\textbf{acc}^{s}+\textbf{acc}^{u})$\cite{xian_zero-shot_2018}. In addition, $\textbf{acc}^{s}_{bi}$ or $\textbf{acc}^{u}_{bi}$ is used to evaluate the seen-unseen binary classification results. We also utilize the proportion $\textbf{P}^{s}=\textbf{acc}_{@1}^{s}/\textbf{acc}_{bi}^{s}$ to evaluate the multi-class classification performance under the correct binary classification.

The implementation details, hyper-parameter sensitivity analysis, and model configuration can be found in Appendix D due to space constraints.

\subsection{Comparison with Baselines}
We compare ZeroDDI with baselines in both CZSL and GZSL scenarios, and the results are shown in Table \ref{baselines}. Overall, ZeroDDI achieves the best performances on all unseen metrics (i.e., $\textbf{acc}^{u}$, $\textbf{P}^{u}$ and $\textbf{acc}^{H}$), which indicates that our method has a superior ability to deal with unseen DDIE prediction problem. Specifically, we have the following observations: (1) Compared with 3DGT-DDI, our method has salient advantages in the ZS-DDIE task, as evidenced by the 9.32\% performance gain on the unseen metrics $\textbf{acc}^{u}_{ave}$ of CZSL and 7.92\% performance gain on $\textbf{acc}^{u}_{ave}$ of GZSL. This not only shows the promising unseen prediction performance of our method but also underscores the necessity of specifically tailoring methods for ZS-DDIE tasks. (2) Comparing the performances of class-based and attribute-based models, we find that, in most cases (i.e., ZSLHinge, ZSLCE, and ZeroDDI), the class-based representations perform better. This demonstrates that, in the ZS-DDIE task, the transferability of class-based representations learned by a PLM may be greater than that of attribute-based representations encoded by the shared attributes among classes. Thus our BRL module is constructed based on the PLM. (3) ZeroDDI outperforms ZeroDDI$^{1}$ and ZeroDDI$^{2}$ demonstrates that the designed BRL, which emphasizes the \emph{Effect} attribute with key biological semantics on the class-based representation, can combine the advantages of both attributes and DDIE textual description for unseen DDIE prediction. (4) Compared with ZSLTriplet$^{1 or2}$, ZeroDDI$^{1 or 2}$ also get competitive advantages, which reveals that our DUA strategy benefits unseen prediction.

\begin{figure}[t]%
\centering
\includegraphics[width=8cm]{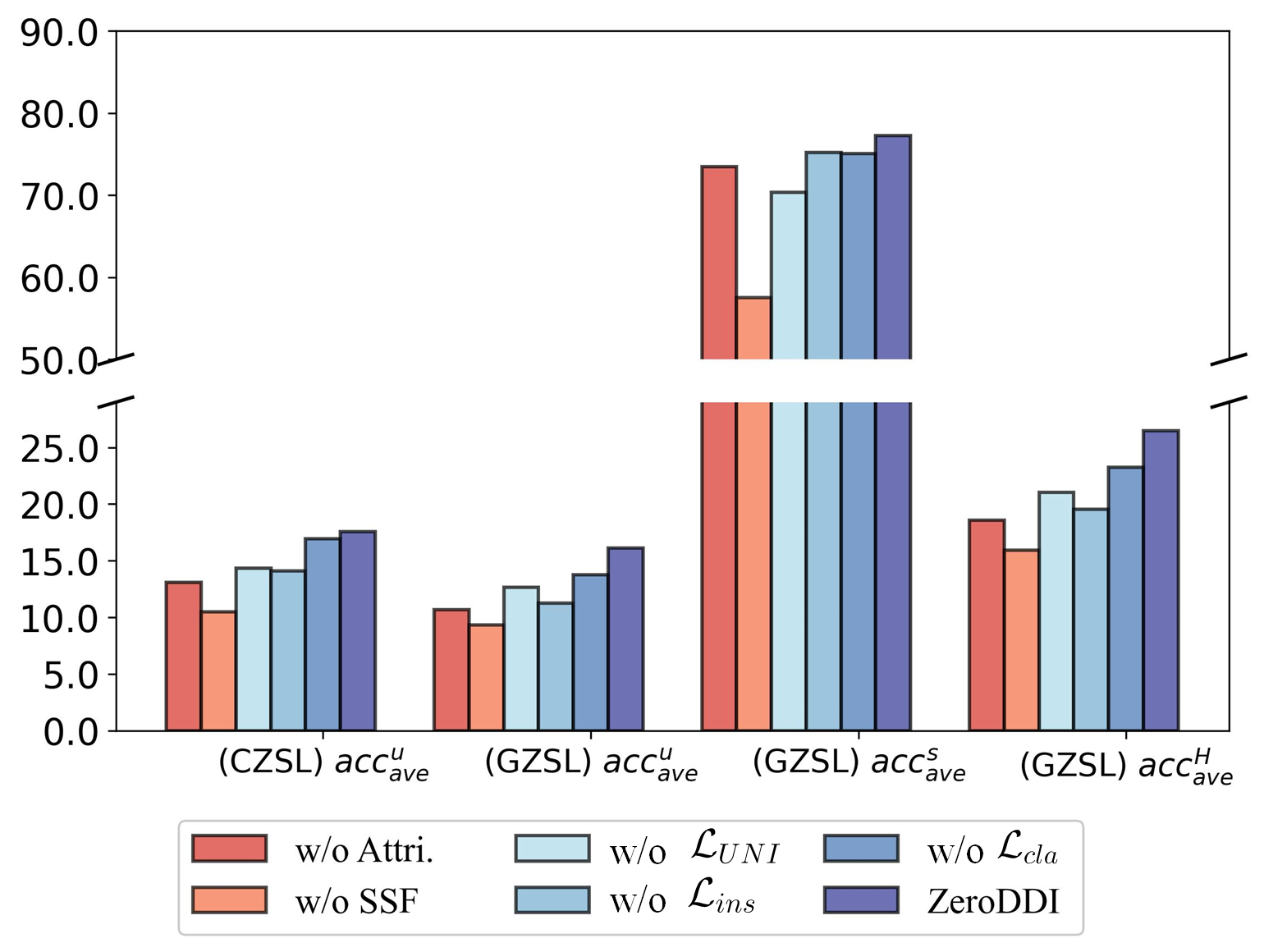}
\caption{Performance (in \%) comparisons of ZeroDDI with its variants in the CZSL and GZSL scenarios.}
\label{xiaorong}
\end{figure}

\subsection{Ablation Study}

We conduct an ablation study to validate the contribution of important components in ZeroDDI, i.e., attribute-level semantics, the SSF component, and dual-modal uniformity loss. The results are shown in Figure \ref{xiaorong}. Specifically, (1) after removing the input of attribute-level semantics (w/o Attri.), the performance drops, which indicates that the attribute-level semantics extracted by our method can improve the performance of the model for the ZS-DDIE task. We will further discuss this in Section 4.4. (2) The model shows a significant decline after replacing the fusion component SSF with the average operation of bi-level token-wise features (w/o SSF), which proves the necessity of selective bi-level fusion and the effectiveness of molecular substructure for distilling discriminative semantic information. (3) Removing the dual-modal uniformity loss (w/o $\mathcal{L}_{UNI}$) or any one part of this loss function (w/o $\mathcal{L}_{cla}$ or w/o $\mathcal{L}_{ins}$) undermines the accuracy, demonstrating that the uniformity loss assist in improve the discriminative ability, and $\mathcal{L}_{cla}$ and $\mathcal{L}_{ins}$ are all useful for ZS-DDI task. We will further discuss this in Section 4.5.

\subsection{Effectiveness of Bio-Enhanced Representation Learning}

\begin{table}[t]
\centering
\setlength{\tabcolsep}{1.5mm}
{
\begin{tabular}{ccccc}
\toprule
\multirow{2}{*}{PLM} &\multirow{2}{*}{\begin{tabular}[c]{@{}c@{}}Attri. Level\\ Semantic\end{tabular}} & \multicolumn{1}{c}{CZSL} & \multicolumn{2}{c}{GZSL} \\  \cmidrule(r){3-3}  \cmidrule(r){4-5} 

& &$\textbf{acc}^{u}_{ave}$    &  $\textbf{acc}_{ave}^{u}$ &$\textbf{acc}_{ave}^{H}$ \\ \midrule

\multirow{2}{*}{SCIBERT} &$\checkmark$& 13.79 & 12.47 &  21.31 \\
 &-&10.84  & 8.58 &14.65    \\ \cmidrule(r){1-5} 

\multirow{2}{*}{PubMedBERT} & $\checkmark$&15.56 & 11.97 &  20.44 \\
 & -&9.60 & 6.94 &  12.40 \\ \cmidrule(r){1-5} 

\multirow{2}{*}{BioBERT} &  $\checkmark$ & \textbf{17.57}   & \textbf{16.13}  &  \textbf{26.48} \\
  & -& 14.66   & 11.72  &  20.21 \\

    \bottomrule
\end{tabular}}
\caption{Performance(in \%) comparisons of ZeroDDI with its variants that replace the BioBERT by different PLM models in CZSL and GZSL scenarios. }
\label{bert}
\end{table}

In this section, we further discuss the effectiveness of two components (attribute-level semantics and SSF) of BRL in unseen DDIE prediction.

\begin{figure}[t]%
\centering
\includegraphics[width=8.5cm]{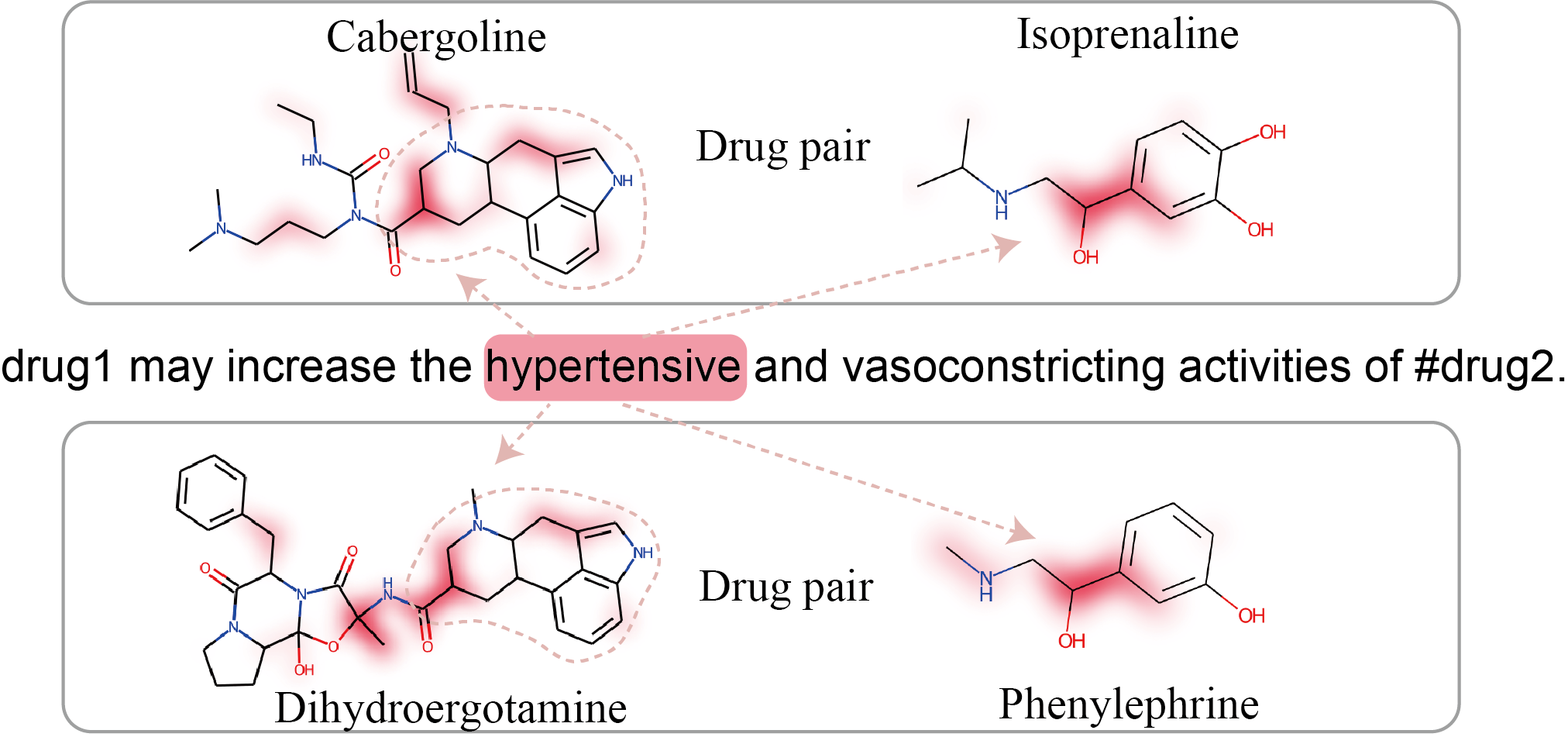}
\caption{The visualization of an example of DDIE textual description with its corresponding drug pair molecular structures. The attention scores between the word "hypertensive" and substructures are highlighted in red colour. }
\label{case}
\end{figure}

For attribute-level semantics, we compare the variants of ZeroDDI (by replacing the BioBERT with SCIBERT \cite{beltagy_scibert_2019} or PubMedBERT \cite{PUBMEDBERT} whose versions are shown in Appendix D.3) equipped with/without the attribute-level semantic information. Table \ref{bert} shows that our attribute-level semantics can provide additional stable gains of transferability, leading to better performance in unseen DDIE prediction no matter which PLM is based. For SSF, we visualize an example to verify that SSF can construct a reasonable relationship between molecular substructures and the bi-level text tokens. We randomly select two drug pairs from the hits of a DDIE in the test set, then visualize them and highlight one word "hypertensive" with its most relevant substructures in Figure \ref{case}. As the website \footnote{\url{https://go.drugbank.com/drug-interaction-checker}} reports that "coadministration with \emph{adrenergic agonists} with hypertensive potential may lead to increased cardiac output and blood pressure due to the vasoconstricting effects of \emph{ergot derivatives}". In the example, Cabergoline and Dihydroergotamine are \emph{ergot derivatives}, and the SSF can highlight the part of their parent structure of ergot derivatives, which proves the effectiveness of the constructed relationship between tokens and substructures and indicate the potential interpretability of SSF.  

\subsection{Effectiveness of Dual-Modal Uniform Distribution
}

\begin{figure}[t]%
\centering
\includegraphics[width=8.7cm]{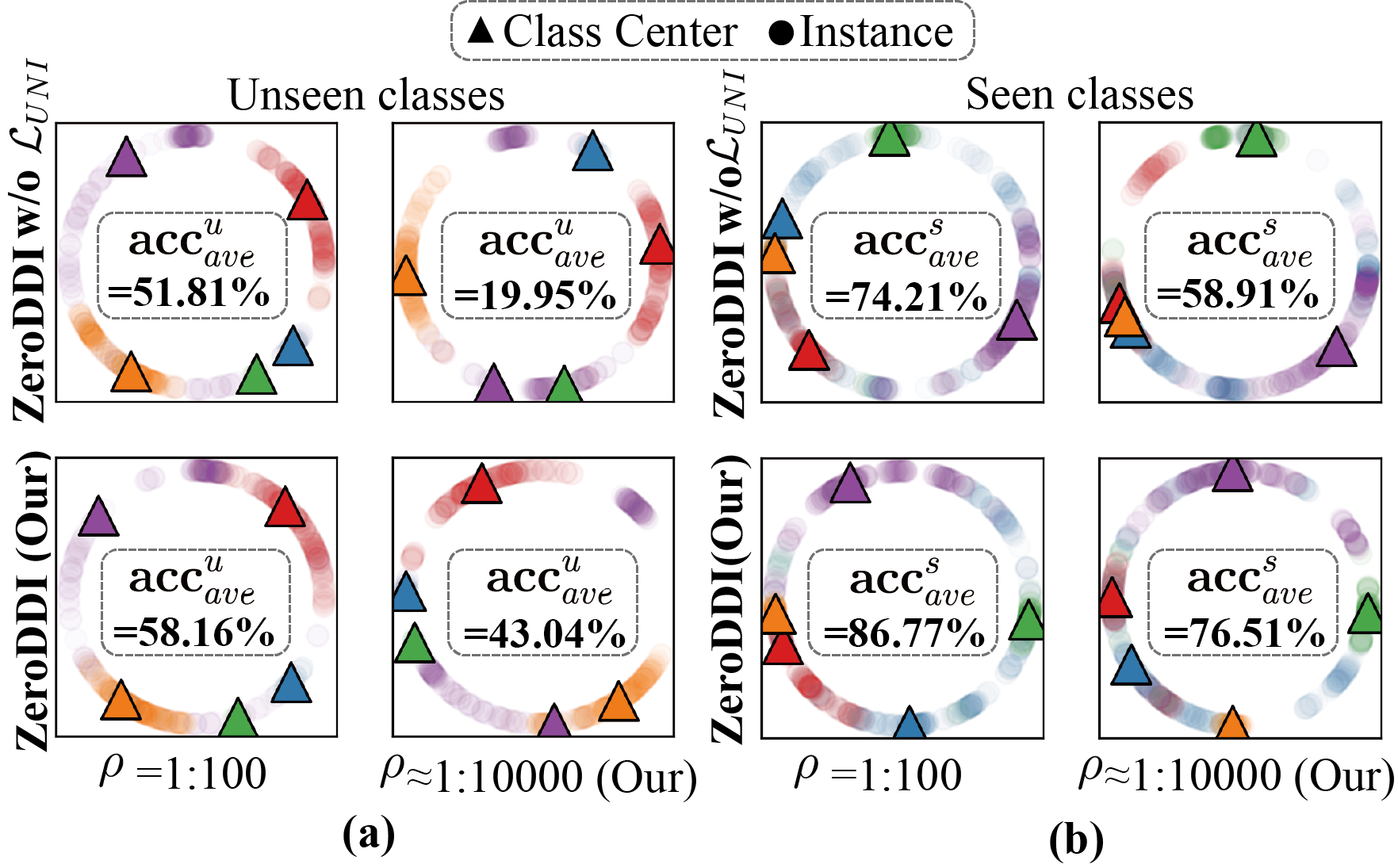}
\caption{The visualization of drug pair representation distribution of test unseen and seen classes in GZSL scenario. Class Center denotes the center of all instances in a class. $\rho$ denotes imbalance ratios of training data. The $\textbf{acc}_{ave}$ here is the average accuracy of five DDIE classes.}
\label{uniform}
\end{figure}

We further discuss the effectiveness of dual-modal uniformity loss for the class imbalance issue in seen and unseen DDIE prediction by comparing ZeroDDI with ZeroDDI (w/o $\mathcal{L}_{UNI}$). To simulate the relatively balanced scenario, we reconstruct the training set with the imbalance ratio $\rho$ is 1:100, where $\rho$ is defined as the number of samples in the most frequent class divided by that of the least frequent class \cite{li_targeted_2022} (here we denote least frequent class has ten instances at least). We randomly select five unseen and seen DDIE classes from the results of the test set, respectively, and visualize them in Figure \ref{uniform}. From the results, we have the following observations: (1) For unseen DDIE prediction, the drop of accuracy from the scenario $\rho =$ 1:100 to 1:10000 shows that the class imbalance has an impact on unseen DDIE prediction to some extent, which demonstrates that the necessity of tackle class unbalance challenge in unseen DDIE prediction. (2) The class imbalance can cause a drop of performance in both seen and unseen classes and lead to unclear and inseparable decision boundaries as shown in the right column of Figure \ref{uniform}(a) and (b), while our method can decelerate this problem. This indicates that $\mathcal{L}_{UNI}$ can improve the discriminability of the model and mitigate the adverse effects of class imbalance.

\subsection{Zero-Shot DDIE Application Analysis}

In this section, we conduct an application analysis to validate the factuality of the zero-shot DDIE setting in this work and evaluate the practical capabilities of ZeroDDI. Firstly, we choose the latest dataset (i.e., DrugBank v5.1.11) as the Novel dataset and take the dataset used in this work as the Existing Dataset. Through the data process for the Novel dataset, we find that, in the Novel Dataset, 6 DDIEs are composed of the attributes from the Existing Dataset but are not included in the Existing Dataset. It not only proves that novel DDIEs are increasing but also proves that novel DDIEs can be composed of existing attributes and further indicates the practical significance of our evaluation setting. Then, we use the Existing Dataset to train the ZeroDDI and use the trained model to predict 6 novel DDIEs in the Novel Dataset. When there are no labelled instances in the training set, ZeroDDI can also achieve 61.11 \% in $\textbf{acc}^{u}_{ave}$, which shows the application ability of our method. More details and experiments are shown in Appendix E.

\section{Conclusion}
This is the first work that pays attention to the zero-shot DDIE prediction and proposes a novel method, called ZeroDDI, to predict zero-shot DDIEs. Specifically, we designed a biological semantic enhanced DDIE representation learning module to learn suitable DDIE representations containing enhanced key biological semantic and substructure-guided discriminative semantics, leading to better knowledge transferability from seen DDIEs to unseen DDIEs. Moreover, we design a dual-modal uniformity alignment strategy to uniform the distribution of drug pair representations and DDIE representations in unit sphere and thus mitigate the class imbalance issue. The extensive experiments show that ZeroDDI can produce superior performance in zero-shot DDIE prediction, our designed module can effectively promote the performances, and ZeroDDI is a promising tool for practice application.

\appendix

\section*{Acknowledgments}
This work was supported by the National Natural Science Foundation of China (62372204, 62072206, 61772381, 62102158); Huazhong Agricultural University Scientific \& Technological Self-innovation Foundation; Fundamental Research Funds for the Central Universities (2662021JC008, 2662022JC004). The funders have no role in study design, data collection, data analysis, data interpretation, or writing of the manuscript.

\section*{Contribution Statement}

Wen Zhang is the corresponding author of this study. Ziyan Wang is responsible for the paper writing, data annotating, experimental code, figure drawing, and experimental design. Zhankun Xiong helped with paper writing, figure drawing, and experimental design, which contributed equally to this paper. Feng Huang provided help in paper writing and Xuan Liu provided help in data annotating. 


\bibliographystyle{named}
\bibliography{ijcai24}

\newpage
\section{Evaluating representation learned from DDIE textual descriptions }

\begin{figure}[h]%
\centering
\includegraphics[width=8.5cm]{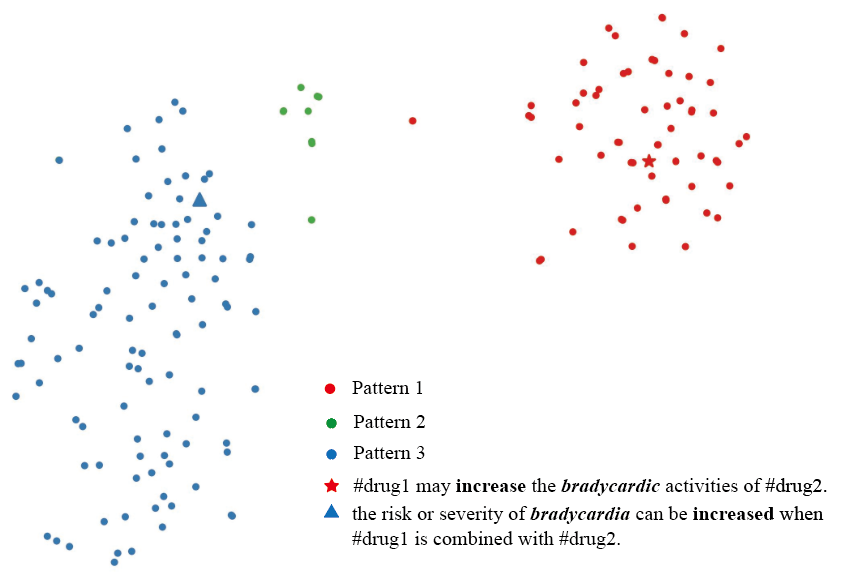}
\caption{Visualization of representations learned by DDIE textual descriptions. Different colours correspond to different \emph{Pattern}.}
\label{dis}
\end{figure}

Before training the model, we evaluate the relatedness of the representations learned from DDIE textual descriptions. Specifically, we first obtain the textual DDIE representations using BioBERT, which is pre-trained on large-scale biomedical corpora. Then we visualize the DDIE representations and attach different colours to DDIEs with different \emph{Pattern} attributes. As shown in Figure \ref{dis}, the representations with different \emph{Pattern} have distinct boundaries, which demonstrates that the \emph{Pattern} in DDIEs has great influences on the semantic relatedness of DDIEs. Moreover, we highlight two examples by using a triangle symbol and a pentagram symbol. Although they have similar biological semantics (such as similar \emph{Effect} attribute and the same \emph{Sign}), they get a large distance in the semantic space due to the different sentence patterns. This indicates that only using DDIE textual description to represent the semantics of DDIE is not reliable and more biological knowledge needs to be emphasized. 

\section{Related works of DDIE Prediction}


\textbf{It is worth noting that} a recent study TextDDI \cite{zhu_learning_2023} has mentioned that they tackle a zero-shot drug-drug interaction task. However, the zero-shot drug-drug interaction event prediction (ZS-DDIE) task proposed in our paper is completely different from theirs. Their "zero-shot" refers to classifying novel drug pairs that are not included in the training set, which is commonly known as an inductive setting in the DDI field. Our "zero-shot" refers to classifying novel drug pairs into \textbf{novel DDIE classes} that are not included in the training set, which is inherited from computer vision. In addition, TextDDI cannot be used to tackle the ZS-DDIE task because they have not encoded the DDIE.

\section{Drug Pair Encoder}

\begin{figure}[h]%
\centering
\includegraphics[width=9cm]{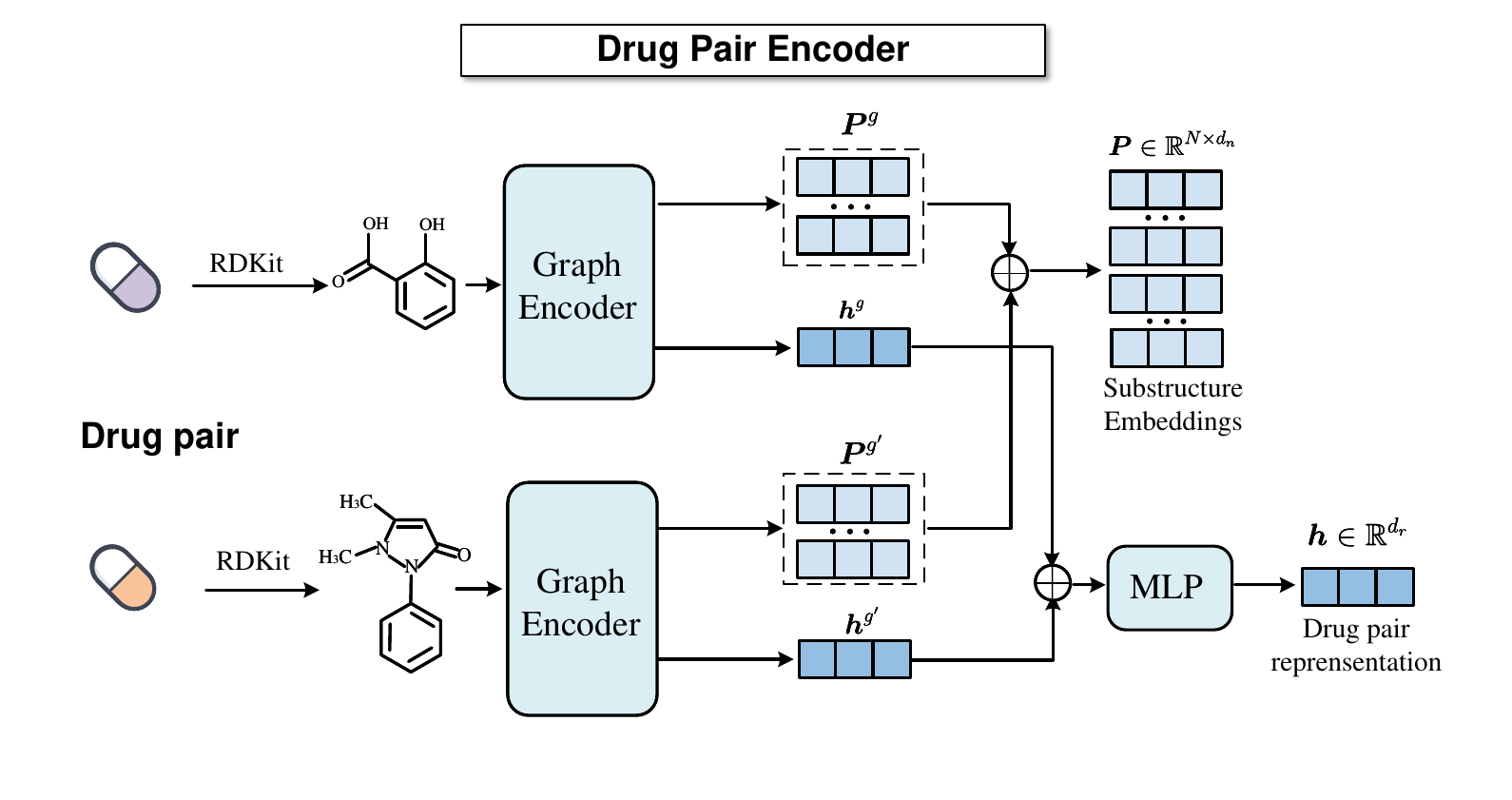}
\caption{The architecture of Drug Pair Encoder.}
\label{drugpair}
\end{figure}

We use a Drug Pair Encoder to learn the drug pair representations $\boldsymbol{h} \in \mathbb{R}^{d_{r}}$ and molecular substructure embeddings $\boldsymbol{P} \in \mathbb{R}^{N \times d_{n}}$, that mentioned in Section 3.3. The process is shown in Figure \ref{drugpair}.

For a drug $g$, we use RDKit to convert it to a molecular graph $\mathcal{G}=(\mathcal{V}, \mathcal{E})$, where $\mathcal{V}$ is the set of nodes representing the atoms and $\mathcal{E} \subset \mathcal{V} \times \mathcal{V}$ is the set of edges representing the bonds. We utilize a Graph Encoder \cite{zhu_molecular_2022} to get the drug embeddings and drug substructure embeddings. In Graph Encoder, firstly Graph Isomorphism Network (GIN) \cite{xu2018how} is used to get the node embeddings $\boldsymbol{X}^{g} \in \mathbb{R}^{|\mathcal{V}| \times d_{v}}$ of a drug. Then the node embeddings are summarized into a graph-level embedding $\boldsymbol{h}^{g}\in \mathbb{R}^{d_{n}}$ through a READOUT function, i.e.,  $\boldsymbol{h}^{g}=\operatorname{READOUT}(\boldsymbol{X})$. Meanwhile, the node embeddings are transformed into substructure embeddings $\boldsymbol{P}^{g} \in \mathbb{R}^{N_{0}\times d_{n}}$ of drug $g$ by a Transformer-like learner, which is formulated by:

\begin{equation}
    \boldsymbol{Q}^{g}=\boldsymbol{Q}_0^{g} \boldsymbol{W}^{g}_Q, \quad \boldsymbol{K}^{g}=\boldsymbol{X}^{g} \boldsymbol{W}_K^{g}, \quad \boldsymbol{V}^{g}=\boldsymbol{X}^{g} \boldsymbol{W}_V^{g}
\end{equation}
\begin{equation}
\boldsymbol{A}^{g}=\operatorname{softmax}\left(\frac{\boldsymbol{Q}^{g} \boldsymbol{K}^{g\top}}{\sqrt{d}}\right)
\end{equation}

\begin{equation}
\boldsymbol{P}^{g}=\operatorname{ReLU}\left((\boldsymbol{Q}^{g}+\boldsymbol{A}^{g} \boldsymbol{V}^{g}) \boldsymbol{W}_P^{g}\right),
\end{equation}
where the $\boldsymbol{Q}_0^{g}\in \mathbb{R}^{N_{0}\times d_{n}}$ is the learnable initialized substructure embeddings aiming to represent the cluster prototypes of different kinds of learned node-centred substructures; $N_{0}$ represents the number of substructures of a drug; $\boldsymbol{W}^{g}_Q$, $\boldsymbol{W}^{g}_K$, $\boldsymbol{W}^{g}_V$ and $\boldsymbol{W}^{g}_P$ are learnable linear transformations. 

For drug pairs, we first use the process above to get the graph-level embedding and substructure embedding of two drugs $g$ and $g^{\prime}$, i.e., $\boldsymbol{h}^{g}$, $\boldsymbol{h}^{g^{\prime}}$, $\boldsymbol{P}^{g}$, $\boldsymbol{P}^{g^{\prime}}$. Then we concatenate $\boldsymbol{P}^{g}$ and $\boldsymbol{P}^{g^{\prime}}$ to obtain the molecular substructure embedding of drug pair $\boldsymbol{P} \in \mathbb{R}^{N\times d_{n}}$, where $N=2\times N_{0}$. Then we concatenate $\boldsymbol{h}^{g}$ and $\boldsymbol{h}^{g^{\prime}}$ and put them into an MLP to obtain the drug pair representation $\boldsymbol{h} \in \mathbb{R}^{d_{r}}$.

\section{Experiment Set}
\subsection{Dataset}

The dataset conducted in this work contains data sourced from the DrugBank v5.1.9 database and the MeSH database. We extract drugs, DDIs, textual descriptions of DDIE, and simplified molecular-input line-entry systems (SMILES) of drugs from DrugBank, and extract the attribute-level textual descriptions from MeSH. The specific process of data processing is as follows:

\textbf{First step: obtain DDIE from raw data.} Every DDI is attached with a textual description in the DrugBank dataset, for example, for drug pair "Valsartan" and "Minoxidil", their textual description is "Valsartan may increase the hypotensive activities of Minoxidil." To apply these to the DDIE classification task, the specific drug names in raw textual descriptions (such as "Valsartan" and "Minoxidil") are replaced with two identifiers (i.e., "\#drug1" and \#drug2"). Thus the final DDIE textual descriptions are changed to "\#Drug1 may increase the hypotensive activities of \#drug2", which can be used to describe a class of a series of drug pairs. We filtered the DDIE textual descriptions that contained more than two drug names.

\textbf{Second step: attach attribute annotations for every DDIE.} For \emph{Effect} attributes, we use the StanfordNLP tool to locate every \emph{Effect} attribute words in DDIE textual descriptions and then find a textual description in MeSH dataset with the closest meanings to every \emph{Effect} words. We ensure that every DDIE has at least one attribute-level description and filtered out the DDIE without attribute-level description. According to this, our dataset has 114 \emph{Effect} attributes. In addition, we attach \emph{Sign} and \emph{Pattern} for every DDIE, the former has "increase" and "decrease" two elements and the latter has three kinds of sentence patterns shown in Figure 1 in the Introduction.

\begin{table*}[t!]
\centering
\small

\begin{tabular}{lll}
\toprule
Hyper-parameter       & Description                                                      & Value  \\ \midrule
The number of layers of GIN  & The number of layers of GIN & 2     \\
$d_{v}$  & the dimension of node embeddings       & 300      \\
$d_{n}$  & the dimension of substructure embeddings       & 300      \\
$d_{r}$ & the dimension of drug pair representations                              & 256     \\
$N$  & the number of substructure representations           & 30   \\
$d_{t}$ & the dimension of the token feature of text & 768\\
learning rate         & learning rate  for Adam optimizer                                 & 0.0001 \\
epoch                 & the number of training epochs                                    & 100    \\
batch size           & the input batch size                                            & 128   \\
    $\tau$      &  temperature coefficient  of contrastive loss                                         & 0.9 \\
    $\lambda$      & hyper-parameter of total loss                                      & 0.7 \\

\bottomrule

\end{tabular}
\caption{The hyper-parameters of ZeroDDI.}
\label{hy}
\end{table*}

\subsection{The Detail of Baselines} 

The ZSL baselines from computer vision we chose in Section 4.1 mainly for two criteria: 1) The image encoder can be easily replaced with the drug pair encoder. 2) For evaluating the transferability of different semantic sources (class-based or attribute-based), we choose the visual ZSL method that does not deeply rely on one specific semantic source.

\textbf{Class-based representation.} The token features learned from BioBERT are the same as those in ZeroDDI. Then the token features are averaged to obtain the class-level text representation.

\textbf{Attribute-based Representation.} We employ one of the commonly used attribute encoding methods, which encodes all attributes as a binary vector. If one attribute is in the current DDIE, the value of this dimension is 1, otherwise, it is 0 \cite{cao_research_2020}. Then we design a 3-layer DNN with a layer normalization to transfer the binary vector to the class attribute representations.

\subsection{Model Configuration}

We develop our model on the machine with a 24 vCPU Intel(R) Xeon(R) Platinum 8352V CPU @ 2.10GHz (CPU) and two NVIDIA GeForce RTX 4090s (GPU). Our model is implemented with PyTorch (1.11), PyTorch-geometric (2.1.0), rdkit (2022.03.3) and deepchem (2.7.1). The versions of BioBERT, SCIBERT, and PubMedBERT are biobert base v1.2, scibert-scivocab cased, and PubMedBERT (abstracts + full text), respectively. Moreover, hyper-parameters of ZeroDDI are shown in Table \ref{hy}. Most basic hyper-parameters of ZeroDDI are set by experiences, such as learning rate, epoch and batch size. The $\tau$ and $\lambda$ are chosen from a predefined set of parameter values, which are shown in Appendix D.4 in detail. 

\begin{figure}[h]%
\centering
\includegraphics[width=7.6cm]{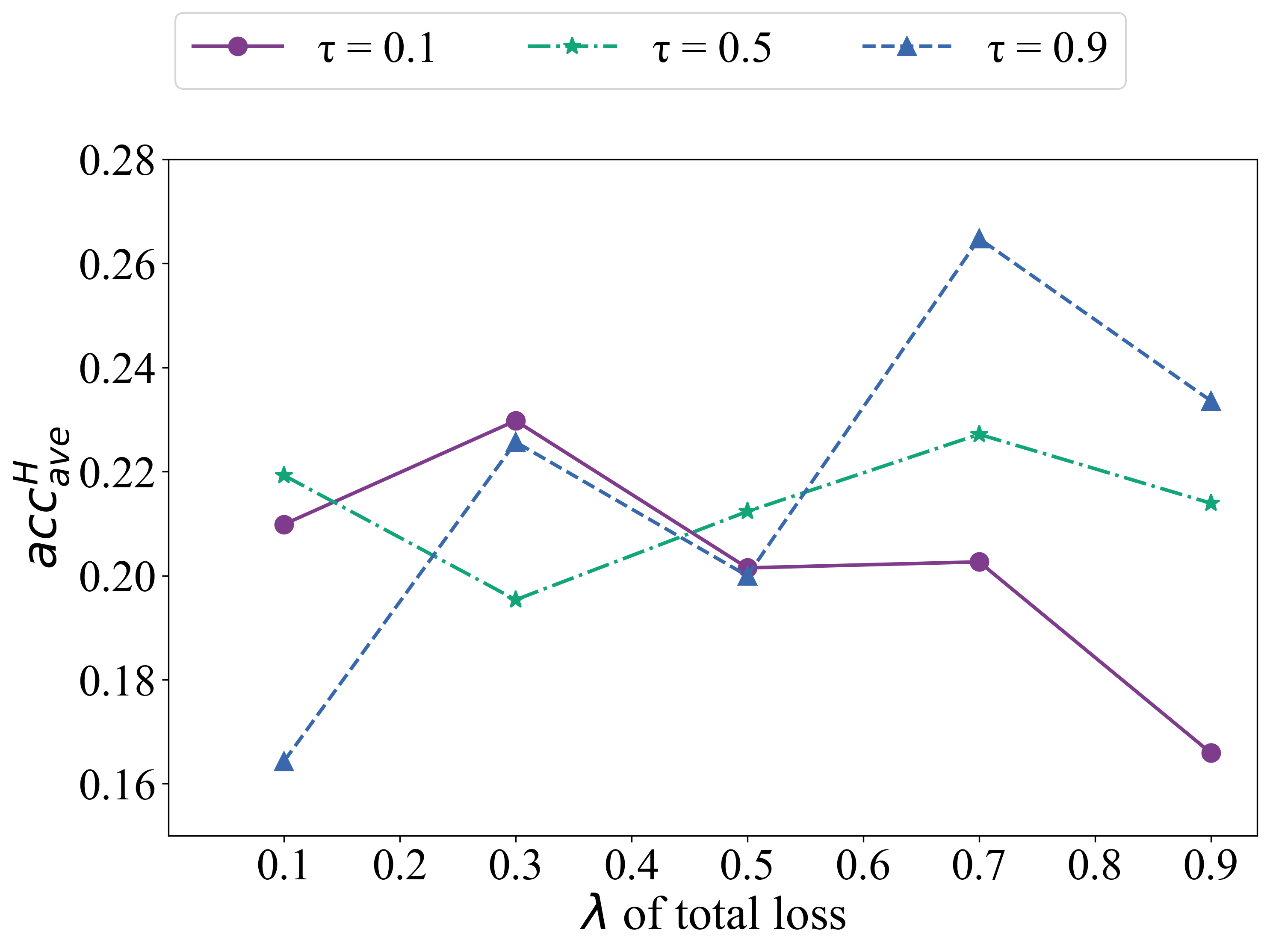}
\caption{Hyper-parameter sensitivity analysis.}
\label{lr}
\end{figure}

\subsection{Hyper-parameter Sensitivity Analysis}
We investigate the sensitivity of our method to these two hyper-parameters: the hyper-parameter of temperature $\tau$ in Eq.(4) and The hyper-parameter of temperature $\lambda$ in Eq.(8). We vary $\lambda$ in the range of  $\{0.1, 0.3, 0.5, 0.7, 0.9\}$ and $\tau$ in the range of $ \{0.1, 0.5, 0.9 \}$. The performance of harmonic mean $\textbf{acc}_{ave}^{H}$ is shown in Figure \ref{lr}. According to Figure \ref{lr}, we find that $\tau$=0.9 and $\lambda$=0.7 achieve the best performance. The performance of our method degrades when the $\lambda$ is too high ($\lambda$=0.9), which demonstrates that the high training weight of uniformity loss will hinder the model's fitting ability and choosing appropriate $\lambda$ is important for improving the performance of ZeroDDI. Moreover, the results show that when the temperature of contrastive loss is small, high $\lambda$ (the importance of uniformity loss) will bring negative effects for ZeroDDI. The reason may be that the contrastive loss with a small temperature will make each instance tend to be more separated, and the instance distribution is likely to be more uniform \cite{wang_understanding_2021}, which has the same objective with uniformity loss. Therefore, the instance is too uniformly distributed in space to reduce the performance of the model and the positive effect of uniformity loss may be weakened when the temperature of contrastive loss is small.

\section{Application Analysis}

In Section 4.6 we have verified the existence of the zero-shot DDIE prediction problem in the real world and the application ability of our proposed model in the scenario that unseen DDIEs are composed of existing attributes. The unseen DDIEs are shown in Table \ref{des}. Since we have proposed ZeroDDI and the simplified version ZeroDDI (w/o Attri.), that is not rely on attribute annotations, for tackling the zero-shot DDIE prediction task. Herein, we also evaluate the application ability of ZeroDDI (w/o Attri.) in a more realistic scenario that unseen DDIEs are compositions of the attributes that include novel \emph{Effect} and \emph{Pattern} attributes. 

\begin{table*}[]
\begin{tabular}{ll|l}
\toprule
  & Existing Dataset                                                                                                                          & Novel Dataset                                                                                                                            \\ \midrule
1 & \begin{tabular}[c]{@{}l@{}}the risk or severity of seizure can be \\ increased when \#drug1 is combined with \#drug2.\end{tabular}        & \begin{tabular}[c]{@{}l@{}}the risk or severity of seizure can be \\ decreased when \#drug1 is combined with \#drug2.\end{tabular}       \\ \midrule
2 & \begin{tabular}[c]{@{}l@{}}the risk or severity of cardiotoxicity can be \\ increased when \#drug1 is combined with \#drug2.\end{tabular} & \begin{tabular}[c]{@{}l@{}}the risk or severity of cardiotoxicity can be \\ decreased when \#drug1 is combined with \#drug2\end{tabular} \\ \midrule
3 & \begin{tabular}[c]{@{}l@{}}\#drug1 may decrease the fluid retaining \\ and vasopressor activities of \#drug2.\end{tabular}                & \begin{tabular}[c]{@{}l@{}}\#drug1 may increase the fluid retaining \\ and vasopressor activities of \#drug2.\end{tabular}               \\ \midrule
4 & \begin{tabular}[c]{@{}l@{}}\#drug1 may increase the \\ vasoconstricting activities of \#drug2.\end{tabular}                               & \begin{tabular}[c]{@{}l@{}}\#drug1 may decrease the \\ vasoconstricting activities of \#drug2\end{tabular}                               \\ \midrule
5 & \#drug1 may increase the diuretic activities of \#drug2.                                                                                  & \#drug1 may decrease the diuretic activities of \#drug2.                                                                                 \\ \midrule
6 & \begin{tabular}[c]{@{}l@{}}the risk or severity of cardiotoxicity can be \\ increased when \#drug1 is combined with \#drug2.\end{tabular} & \#drug1 may increase the cardiotoxic activities of \#drug2.                                                                              \\ \bottomrule
\end{tabular}
\caption{Examples of the novel increased DDIE textual descriptions in Novel Dataset and its similar DDIE textual descriptions in Existing Dataset. Novel Dataset is DrugBank v5.1.11 released on 2024-01-03.}
\label{des}
\end{table*}

We take the Existing Dataset as the training set (seen data), and evaluate the performance of ZeroDDI (w/o Attri.) on the Novel Dataset that exclude the seen instances, seen DDIEs and 6 DDIEs mentioned in Section 4.6. After the process, we get 80 unseen DDIEs with total 206996 drug pairs. In this case, the model without transferability can achieve a base result $\frac{1}{80}$ in $\textbf{acc}^{u}_{ave}$, while ZeroDDI (w/o Attri.) achieves 10.97\% in $\textbf{acc}^{u}_{ave}$, which verify the transferability of our model.

\end{document}